\documentclass[journal,twoside,web]{ieeecolor}
\usepackage{generic}

\makeatletter
\let\NAT@parse\undefined
\makeatother

\usepackage{graphicx}
\usepackage{picinpar}
\usepackage{amsmath}
\usepackage{url}
\usepackage{flushend}
\usepackage[latin1]{inputenc}
\usepackage{colortbl}
\usepackage{soul}
\usepackage{multirow}
\usepackage{array}
\usepackage{pifont}
\usepackage{color}
\usepackage{alltt}
\usepackage{makecell}
\usepackage[hidelinks]{hyperref}
\usepackage{enumerate}
\usepackage{siunitx}
\usepackage{threeparttable}
\usepackage{epstopdf}
\usepackage{pbox}
\usepackage{amssymb}
\usepackage{subfigure}
\usepackage{pifont}
\usepackage{tasks}
\usepackage[ruled]{algorithm2e}
\usepackage{booktabs}
\usepackage[numbers,sort&compress]{natbib}
\usepackage[switch]{lineno}

\begin{document}

\title{Efficient Visual Fault Detection for Freight Train via Neural Architecture Search with Data Volume Robustness}

\author{
	\vskip 0.5em
    Yang~Zhang, Mingying~Li, Huilin~Pan, Moyun~Liu, and Yang~Zhou
\thanks{Manuscript received XX XX, 2023; revised xx xx, 2024; accepted xx xx, 2024.}
\thanks{Y. Zhang, M. Li, H. Pan, and Y. Zhou are with the School of Mechanical Engineering, Hubei University of Technology, Wuhan 430068, China (e-mail: yzhangcst@hbut.edu.cn; 102200036@hbut.edu.cn; hlp@hbut.edu.cn; z-g99@hbut.edu.cn).}
\thanks{M. Liu is with the School of Mechanical Science and Engineering, Huazhong University of Science and Technology, Wuhan 430074, China (e-mail: lmomoy@hust.edu.cn).}
\thanks{Y. Zhang is also with the National Key Laboratory for Novel Software Technology, Nanjing University, Nanjing 210023, China. (e-mail:	yzhangcst@smail.nju.edu.cn).}
}

\maketitle

\begin{abstract}
Deep learning-based fault detection methods have achieved significant success. 
In visual fault detection of freight trains, there exists a large characteristic difference between inter-class components (scale variance) but intra-class on the contrary, which entails scale-awareness for detectors. Moreover, the design of task-specific networks heavily relies on human expertise. As a consequence, neural architecture search (NAS) that automates the model design process gains considerable attention because of its promising performance. However, NAS is computationally intensive due to the large search space and huge data volume. In this work, we propose an efficient NAS-based framework for visual fault detection of freight trains to search for the task-specific detection head with capacities of multi-scale representation. First, we design a scale-aware search space for discovering an effective receptive field in the head. Second, we explore the robustness of data volume to reduce search costs based on the specifically designed search space, and a novel sharing strategy is proposed to reduce memory and further improve search efficiency. Extensive experimental results demonstrate the effectiveness of our method with data volume robustness, which achieves 46.8 and 47.9 mAP on the Bottom View and Side View datasets, respectively. Our framework outperforms the state-of-the-art approaches and linearly decreases the search costs with reduced data volumes.

\end{abstract}

\begin{IEEEkeywords}
Fault detection, Freight trains, Neural architecture search, Date volume robustness, Sharing.
\end{IEEEkeywords}

\section{Introduction}
\IEEEPARstart{F}{ault} detection of freight trains is vital for the safety of railway systems, which provides maintenance decisions by recognizing the defects existing in moving trains. However, traditional manual inspection is inefficient and time-consuming, unsatisfying the safety of freight trains in high-frequency operation. Recently, deep learning~\cite{butera2021precise,zhang2022visual,saufi2020gearbox} has made enormous strides in computer vision, and many researchers work on visual fault detection of freight trains~\cite{2.A.2, 2.A.4, 2.A.5}. Although automated fault detection achieves high accuracy, designing the neural network requires considerable time.  Moreover, different train components remain large-scale variations that entail scale-awareness for detectors, especially one-stage detectors, which need a lot of trial and error to fit our data characteristics.

\begin{figure}[!t] \centering   
\subfigure[Ground-truth] {
\label{fig1:a}     
\includegraphics[width=0.24\columnwidth]{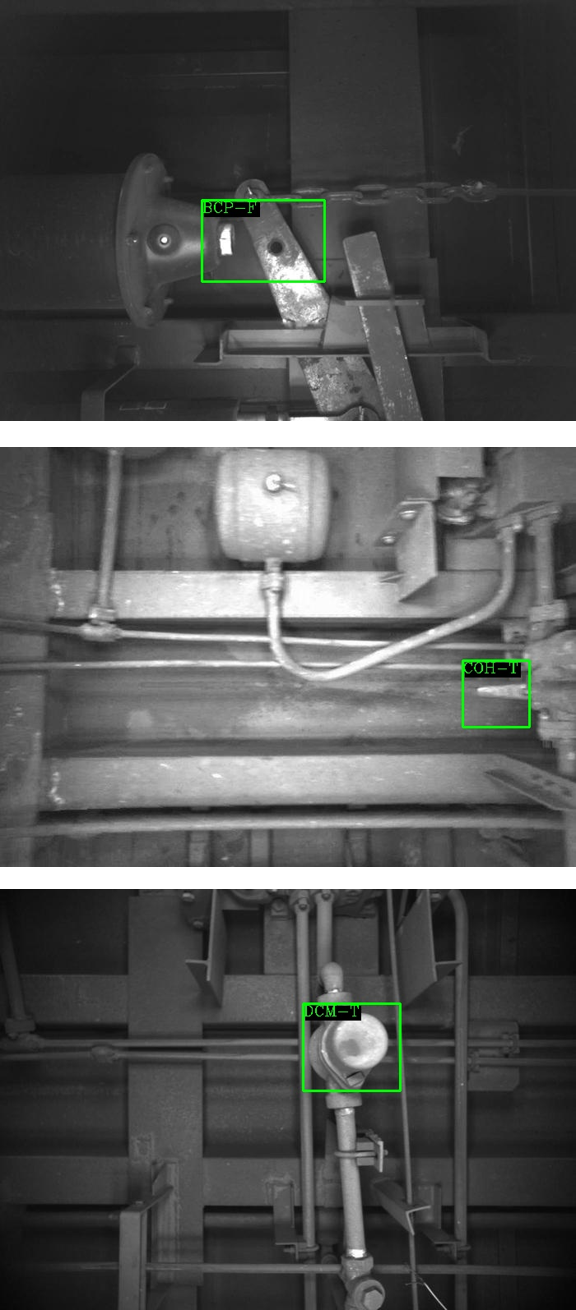}  
} \hspace{-1.3em}
\subfigure[Original] {
 \label{fig1:b}     
\includegraphics[width=0.24\columnwidth]{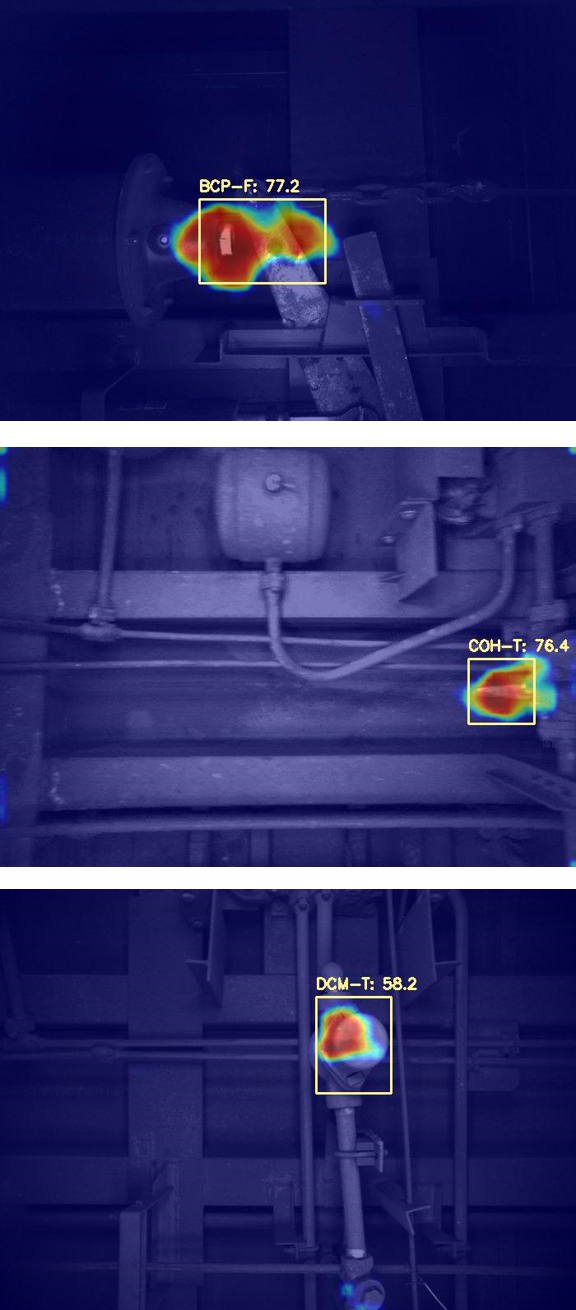}  
} \hspace{-1.3em}  
\subfigure[Ratio 1/2] { 
\label{fig1:c}     
\includegraphics[width=0.24\columnwidth]{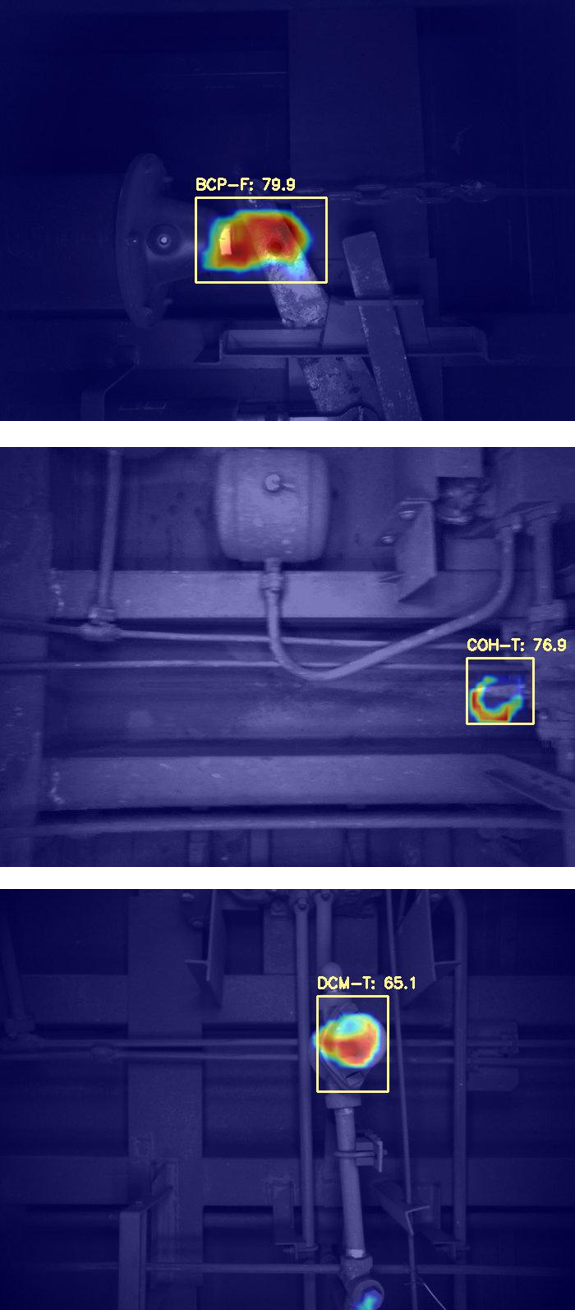}     
} \hspace{-1.3em}   
\subfigure[Ratio 1/4] { 
\label{fig1:d}     
\includegraphics[width=0.24\columnwidth]{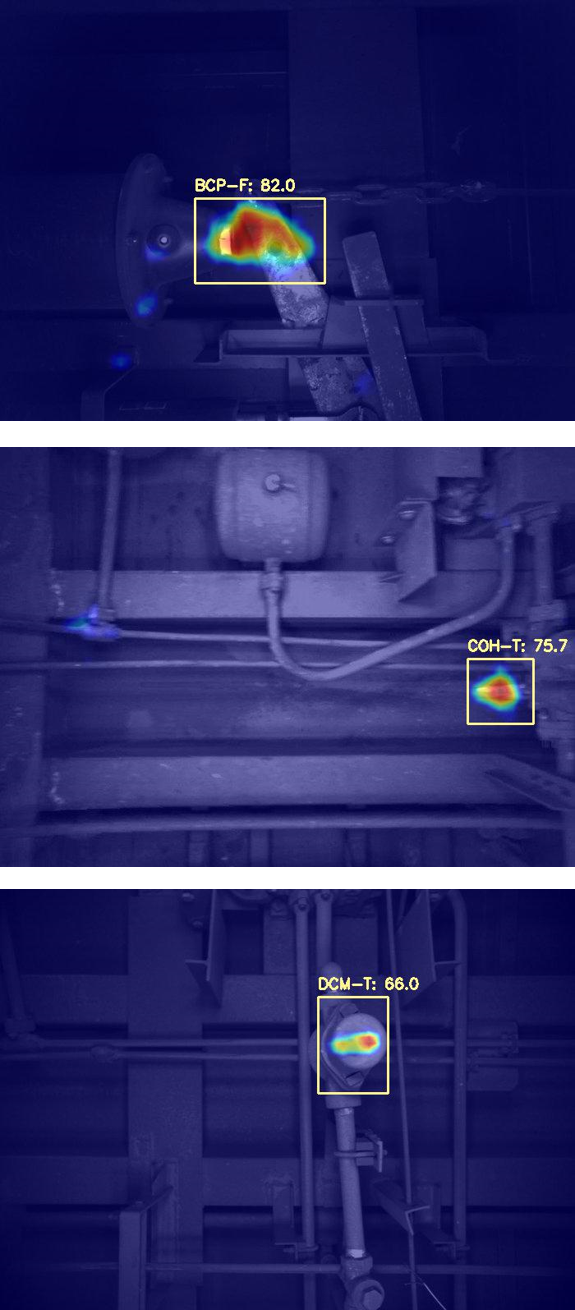}     
}   
\caption{Visual results of models searched on the Bottom View datasets with various volumes. All models exhibit competitive responses in the class response maps and achieve precise localization.}
\label{grad_cam}
\end{figure}

There has been a growing interest in neural architecture search (NAS) that automates the design process of neural networks, which further shows comparative or even better results compared with hand-crafted networks. Many search strategies like reinforcement learning (RL), evolutionary algorithm (EA), and gradient-based (GD) methods are used in NAS, where the search cost of RL- and EA-based methods~\cite{NAS-FPN, NAS-FCOS, DetNAS,2021OPANAS} is considerable, as shown in Table~\ref{search method}. The high search cost is the main problem preventing NAS from applying to the real world. It is usually caused by the search space that is set large to discover optimal designs and adapt different tasks, while the large search space also gives rise to the memory explosion challenge. This problem becomes even worse when using a large-scale dataset. Nevertheless, NAS sometimes performs better on the sub-dataset than on the whole dataset. Inspired by this, we intend to analyze the effect of data volumes on the search process and explore data volume robustness that maintains comparative performance when using reduced datasets to improve search efficiency. When applying NAS to fault detection of freight trains, some challenges are as follows:

\begin{enumerate}[1)]
    \item \textbf{Design of task-specific search space.} Search space has a critical impact on the final performance. Different vision tasks are sensitive to different search spaces, so it is important to focus on the specific task requirements. Therefore, the design of search space needs to carefully consider the challenges in fault detection of freight trains. 
    \item \textbf{Faults in compact intra-class and dispersed inter-class gaps.} The faults of the same regions are similar, while there is a large gap between different kinds of parts. The NAS methods for fault detection of freight trains should consider the data characteristics in order to guarantee the discovery of the best networks in fault detection tasks.
    \item \textbf{Significant computational costs and memory usage.} When directly searching on a large search space or huge dataset, the corresponding memory and computational overhead are so heavy that ordinary hardware cannot support the search process efficiently. It is essential to design an acceleration method for speeding up the search process.\end{enumerate}

\begin{table}[!t]
\renewcommand{\arraystretch}{1.2}
	\caption{Comparison with different NAS-based methods for fault detection of freight trains on bottom view dataset.
	\label{search method}}
	\centering
	\small
	\setlength{\tabcolsep}{1.2mm}{
		\begin{tabular}{lccccc}
			\toprule
                \multirow{2}{*}{\textbf{Method}}&
                \multirow{2}{*}{\textbf{\makecell[c]{Search\\Method}}}&
			\multirow{2}{*}
            {\textbf{AP$_{50}$}} & 
			\multirow{2}{*}
            {\textbf{AR$_{1}$}}&
               \multirow{2}{*}{\textbf{\makecell[c]{Search Cost\\(GPU-days)}}}&
                \multirow{2}{*}
            {\textbf{\makecell[c]{Model\\Size (MB)}}}\\
            & & & & & \\
			\midrule
            NAS-FPN \cite{NAS-FPN}  
            &RL &67.7 &45.3 &\textgreater{100} &548.0  \\
			NAS-FCOS \cite{NAS-FCOS}   
            &RL &69.6 &60.0 &23 &310.3  \\
			DetNAS \cite{DetNAS}   
            &EA &67.4 &57.2 &37 &162.1  \\
            OPANAS \cite{2021OPANAS}
            &EA &70.6 &59.4 &4 &284.3  \\
            NAS FTI-FDet
            &GD &71.6 &60.7 &0.01 &49.0  \\
			\bottomrule
	\end{tabular}}
\end{table}

To this end, we propose an efficient NAS-based framework for fault detection of freight train images (NAS FTI-FDet), aiming to search for a task-specific detection head with capacities of multi-scale representation. 
More receptive field (RF) candidates and convolution types are firstly designed to build scale-aware search space, allowing the searchable head to quickly adapt to data characteristics with better classification and localization. Inspired by the phenomenon that NAS sometimes obtains better performance on sub-datasets instead of the whole datasets as shown in Fig.~\ref{grad_cam}, we then seek an approach for reducing data size to improve search efficiency and further analyze the effect of the data size on the search process to explore data volume robustness based on the designed search space. Finally, to remedy the huge GPU memory caused by the large search space, we introduce an innovative scheme in which each intermediate representation of candidate operations with identical RF share parameters and computations with each other. Such scheme reduces the intermediate tensor to be stored and further speeds up the search process.
Extensive experiments on two typical fault datasets show that our NAS FTI-FDet achieves competitive accuracy compared with the state-of-the-art (SOTA) detectors, \textit{i.e.}, 46.8 and 47.9 mAP, respectively. 
Moreover, our proposed method shows data volume robustness that maintains comparable or even better accuracy with reduced data volumes, linearly decreasing the search cost.

To summarize, our main contributions are as follows:
\begin{enumerate}[1)]
    \item A scale-aware search space is carefully designed with more RF to adapt to data characteristics and enables the searchable head to build a strong multi-scale representation. Then, an innovative method by sharing intermediate computation is proposed to remedy the GPU memory caused by the large search space.
    \item We explore the effect of data size on the search process and enable the data volume robustness based on the specifically designed search space to improve search efficiency using reduced data volumes, which ensures the performance and fast iteration of the models when capturing vast amounts of new data.
    \item We demonstrate that the search time has a linear reduction by using the decreased dataset, accompanied by comparable or better architecture designs compared with the one searched on the full dataset. These findings are also appropriate for some industrial scenarios with scarce data.
    \item Quantitative experiments on two typical fault datasets indicate that our proposed efficient NAS-based framework achieves superior performance compared with the SOTA methods for fault detection of freight train images.
\end{enumerate}

The remainder of this paper is organized as follows. Section~\ref{sec:related_work} introduces related work, and Section~\ref{sec:proposed method} describes the proposed NAS FTI-FDet framework. Section~\ref{sec:experiments} presents the experimental datasets, results, and analysis. Finally, Section~\ref{sec:conclusion} concludes the work and plans for the future.

\begin{figure*}[!t]
    \centering
    \includegraphics[width=6in]{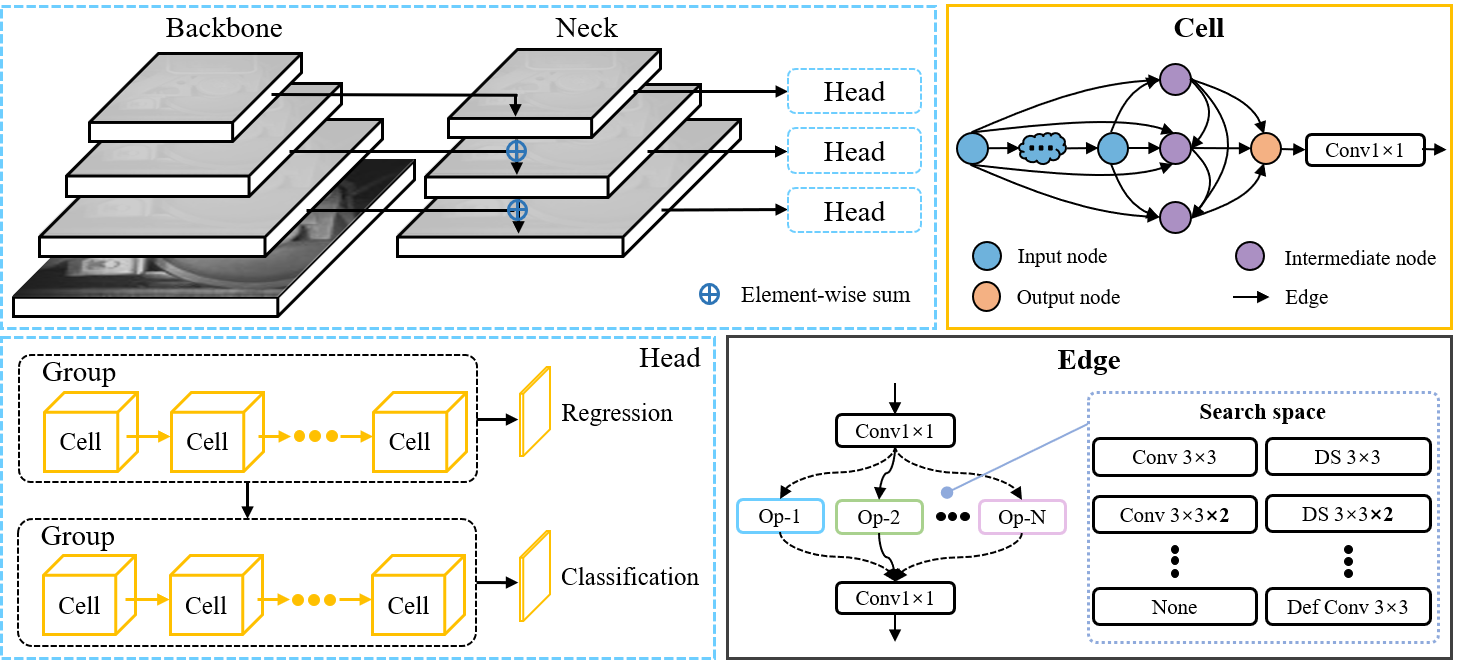}
    \caption{An overview of our proposed search framework for visual fault detection of freight trains. Our method focuses on searching for the optimal head of detectors. The searchable head is constructed by two groups of cells. The edge linking nodes within a cell are composed of two 1$\times$1 convolutions and a search space between the two. The search space contains multiple operations, allowing the edge to search for proper receptive field combinations. Both cell structures and operations on the edges are searchable.
    } 
	\label{framework}
\end{figure*}

\section{Related Work}\label{sec:related_work}
\subsection{Fault Detection for Freight Train Images}
Recently, more researchers have focused on fault detection of freight train images. For example, Sun \textit{et al.}~\cite{2.A.2} proposed a method based on binocular vision to detect bolt-loosening on freight trains. Sun~\textit{et al.}~\cite{2.A.3} proposed to diagnose bogie faults using singular spectrum analysis. These traditional approaches are computationally expensive and cannot satisfy the real-time demand, owing to the task lying in the resource-constrained condition. Moreover, the application of deep learning methods brings new ideas for freight fault detection. Chen~\textit{et al.}~\cite{2.A.4} introduced an automated visual inspection framework that explored structural knowledge of train components for automated visual inspection of trains. Zhou~\textit{et al.}~\cite{2.A.5} utilized a fault detector based on NanoDet-ResNet101 to detect height valves with 97\% accuracy. 
However, these methods are designed for some specific freight components, which greatly affect effectiveness and generality.

\subsection{General Object Detection}
Based on the usage of a region proposal network (RPN), object detection methods can be broadly classified into two streams: one-stage and two-stage detectors.
For one-stage detectors, including CenterNet~\cite{Centernet}, CentripetalNet~\cite{CentripetalNet}, fully convolutional one-stage (FCOS)~\cite{FCOS}, generalized focal loss (GFL)~\cite{GFL}, RetinaNet~\cite{Retinanet}, you only look one-level feature (YOLOF)~\cite{YOLOF}, YOLOX~\cite{YOLOX}, YOLOv7~\cite{wang2023yolov7} and YOLOv8~\cite{yolov8}, these approaches strive to enhance their efficiency by directly classifying predefined anchors without requiring proposal generation. Nonetheless, they may have some accuracy sacrifice due to scale variation problems. Furthermore, two-stage detectors, such as Faster R-CNN~\cite{Faster_R-CNN}, Libra R-CNN~\cite{Libra_R-cnn}, and Sparse R-CNN~\cite{Sparse_R-CNN}, require RPN to generate region of interest (RoI) proposals first and refine them using two subnets. In contrast to one-stage detectors, two-stage detectors demonstrate superior robustness and performance. However, their inference process is less efficient. Moreover, the knowledge distillation~\cite{yang2023skill,huang2022feature} methods are used to transfer knowledge from cumbersome teacher models to improve the performance of lightweight student models for satisfying real-time object detection. Apart from the above methods, a new paradigm transformer has deeply affected the deep learning field. Deformable DETR~\cite{Deformable_DETR} was proposed to solve the slow convergence and deficit of processing image feature maps. Compared with traditional detectors, transformers effectively overcome the difficulties in multi-scale features.

\subsection{Neural Architecture Search for Object Detection}
Existing NAS methods have been growing in popularity and achieved remarkable performance in object detection, which focuses on searching different components of detectors. Ghiasi \textit{et al.}~\cite{NAS-FPN} employed the RL-based method to search for an improved feature pyramid network (FPN).
OPANAS~\cite{2021OPANAS} developed various information paths constituting a novel search space to find the effective FPN architecture with the EA-based method. Aside from FPN, the NAS-FCOS~\cite{NAS-FCOS} searched the detection head and an index within the head indicating where to share weights. Moreover, inspired by the one-shot method, DetNAS~\cite{DetNAS} decoupled the weight training and architecture search to form a backbone in object detectors. However, these NAS-based methods are effective but time-consuming, preventing NAS from being widely applied. To address this issue, 
Auto-FPN~\cite{Auto-FPN} resorted to differentiable NAS pipelines using gradient-based optimization for finding a better FPN and detection head. FAD~\cite{FAD} explored the search space on the head sub-structure and employed a novel search algorithm to speed up the search process.

By contrast, we intend to search for a task-specific detection head in fault detection of freight trains. We carefully design search space and explore the data volume robustness to improve performance and search efficiency. In addition, a novel method is proposed to relieve memory usage.

\section{Proposed Method}\label{sec:proposed method}
In this section, we first introduce the overall framework. Then, we elaborate on how to design a search space suitable for our task scenarios, and a more efficient search scheme is proposed to improve computation. Finally, we present the optimization process.

\subsection{Overall Framework} 
We proposed an efficient NAS FTI-FDet framework based on FCOS by decreasing the data volumes to reduce the search time. Our proposed framework aims at automatically designing the detection head with the capacity of multi-scale representation, coping with a wide range of scale variation problems in our task scenario, as shown in Fig.~\ref{framework}. 

Taking images as input, the backbone performs feature extraction at different scales, and the neck (\textit{i.e.}, FPN) remains the same as the backbone. Then, each level of the FPN is connected to the head to be searched, and weights for the head are shared across different levels.
The searched head is constructed by the two groups of cells, where only one type of cell resides in each group, and the cell structure varies with different groups. Among them, the first group performs bounding box prediction, and the second is for classification. The purpose of this design is that the classification needs more comprehensive information, while the localization prefers detailed location-sensitive feature information.
For the search space, we elaborately develop a search space containing more available RF combinations, which enable the detection head to find a multi-scale effective RF. Furthermore, we present a novel sharing method to overcome the huge GPU memory usage caused by large search space.

\subsection{Design of Search Space} 
Search space design has a critical impact on the final performance of the NAS-based algorithms. As illustrated in Fig.~\ref{framework}, the network is stacked with $L$ cells consisting of $N$ nodes and corresponding edges $E$. The cell is a directed acyclic graph (DAG) $\mathcal{G}$, and the node indicates latent representations, including input, intermediate, and output nodes. Each edge denotes a searched operation and contains two $1\times1$ convolutions $f$ and an operation space \begin{math} \mathcal{O} \end{math} between the two. Let a pair of intermediate nodes be $(i, j)$, where $0 \leqslant i $\textless$ j \leqslant {N-1}$, the $j$-th nodes is thus formulated as:
\begin{equation}\label{equ1}
x_j = \sum_{i<N,o \in \mathcal{O}}^N f_{1}^{c_1,c_2}(o_{i,j}(f_{2}^{c_2,c_1}(x_i)) ),
\end{equation}
where $x_j$ is linked each previous node $x_i$ with corresponding edge. The $f_{1}^{c_1,c_2}$ and $f_{2}^{c_2,c_1}$ are two $1\times1$ convolutions used to maintain the same channel size for all nodes. The $o_{i,j}$ is an operation sampled from the $\mathcal{O}$. To ensure the scalability of the searched architecture, a $1\times1$ convolution is applied on the output node to maintain the same channels among all cells.

The original sub-networks in FCOS consist of only Conv $3 \times 3$. This pattern restricts the RF in the deep stage, which has an impact on the low-resolution features to capture high-level semantics. Additionally, the large characteristic difference between inter-classes in our task entails comprehensive and global representation.
Therefore, we design the $\mathcal{O}$ from the perspective of RF to seek a heterogeneous convolutions mode within the head to capture the effective multi-scale features. The $\mathcal{O}$ is divided into two groups, standard groups utilizing only standard convolution and extended groups that involve special operations such as depth-wise separable (DS), dilated (Dil), and deformable convolution (Def Conv). We apply Dil to enlarge the RF without additional cost and DS to reduce computation. To enable the decoder to learn the sampling location adaptively, we include Def Conv $3 \times 3$, and a no connection (None) operation is introduced to indicate a lack of connection between two nodes. We consider $I=3$ intermediate nodes (layers) in each cell. Each layer has possible RF $R=\{r_1,r_2,...,r_c\}$, and the final combination of RF is denoted as $C=\{c_1,c_2,...,c_l\}$, where  $c_l \in R$. There are $R^I$ possible combinations of RF available, and we aim to search for an effective RF combination to fit our data characteristics, which are defined as:
\begin{equation}\label{equ2}
C_o = argmax\ g(c_l|V),
\end{equation}
where $C_o$ is the searched optimal RF combination, and $g(\cdot)$ denotes performance on the validation dataset $V$.

\begin{table}[!t]
\renewcommand{\arraystretch}{1}
	\caption{Operation used in the search process, $r=2$ denotes dilation rate 2 in dilated convolution.
		\label{operation space}}
	\centering
		\small
		\setlength{\tabcolsep}{1.2mm}{
		\begin{tabular}{ll}
		\toprule
		\textbf{Standard Groups:} & \\
            \midrule
            Op$_1$ : Conv $3\times3$ & Op$_2$ : Conv $3\times3$ + Dil $3\times3$ ($r=2$)\\
            Op$_3$ : Conv $3\times3$ $\times\textbf{2}$ & Op$_4$ : Conv $3\times3$ $\times\textbf{2}$ + Dil $3\times3$ ($r=2$)\\
            Op$_5$ : Conv $3\times3$ $\times\textbf{3}$ \\
            \midrule
            \textbf{Extended Groups:} & \\
            \midrule
            Op$_6$ : DS $3\times3$ & Op$_7$ : DS $3\times3$ + Dil $3\times3$ ($r=2$)\\
            Op$_8$ : DS $3\times3$ $\times\textbf{2}$ & Op$_9$ : DS $3\times3$ $\times\textbf{2}$ + Dil $3\times3$ ($r=2$)\\
            Op$_{10}$: DS $3\times3$ $\times\textbf{3}$ & Op$_{11}$: None  \\ 
            \multicolumn{2}{l}{Op$_{12}$: Def Conv $3\times3$}  \\
                \bottomrule
	\end{tabular}}
\end{table}

Moreover, all operations are decomposed into $3 \times 3$ filters, and more details can be found in Section \ref{fast}. The whole \begin{math} \mathcal{O} \end{math} consists of $K=12$ candidate operations which are listed in Table \ref{operation space}. 
In DAG, each cell has $E_I=2+3+4=9$ mixed edges for intermediate nodes based on the connection mode as previously mentioned, and we only select one operator out of $K$ during the search process. The total architecture combinations in our search space are calculated as $12^{9}=5.16 \times 10^{9}$. So, the complexity of our proposed method is $\mathcal{O}(K^{E_{I}})$. The huge search space requires a formidable computational burden, which necessitates the acceleration method for the search process and will be introduced in the following section.

\subsection{Representation Sharing}
\label{fast}

As discussed before, large search space causes formidable computational and memory burdens. We present a novel sharing method to solve this problem.
We initially introduce filter decomposition as proposed in~\cite{2014Very} to optimize computation. Convolutions with larger spatial filters can be decomposed into some $3 \times 3$ convolutional layers, which not only decreases the number of parameters but also improves non-linear capability. For instance, the Conv $5 \times 5$ can be substituted by a sequence of two $3 \times 3$ convolutional layers. 
Let the weights of kernel and its corresponding decomposed kernels as $\textbf{w}_{k \times k}$ and $\textbf{w}_{3 \times 3}^{km}$, the weight of Conv $5 \times 5$ is product of two decomposed kernel weights, \textit{i.e.,} $\textbf{w}_{5 \times 5} = \textbf{w}_{3 \times 3}^{51} \cdot \textbf{w}_{3 \times 3}^{52}$, and all factorized operations can be denoted as follows:
\begin{equation}\label{equ3}
\begin{aligned}
&\textbf{w}_{k \times k} = \prod_{m<M,}^M \textbf{w}_{3 \times 3}^{km},
\end{aligned}
\end{equation}
where $M$ is the number of decomposed kernels.

\begin{figure}[!t]
    \centering	
    \includegraphics[width=3.2in]{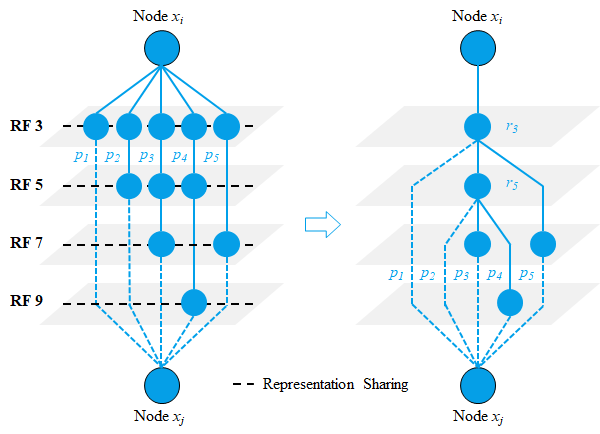}
    \caption{Representation sharing. Each $p_{i}$ indicates a path in which the solid line represents an operation. Each sphere lying path denotes intermediate representations. Large filters are firstly factorized into many 3 $\times$ 3 filters, and then those intermediate representations at the same RF level are shared.}
    \label{representation sharing}
\end{figure}

Once the candidate operations are factorized, there are more intermediate representations (feature maps) with different RF for each factorized operation. As for the same type of convolution, intermediate representations with the same RF are similar. So, we can reuse the intermediate representations at the same RF level for each operation to reduce memory footprint. Moreover, given the similarity between intra-classes, extra parameters may create feature redundancy. We can take parameter sharing across decomposed convolutional layers for one operation. To illustrate the sharing process, we show the case of Conv $5 \times 5$ and Conv $7 \times 7$ as follows: 
\begin{equation}\label{equ4}
\begin{aligned}
\textbf{w}_{5 \times 5} &= \textbf{w}_{3 \times 3}^{51} \cdot \textbf{w}_{3 \times 3}^{52}, \\
\textbf{w}_{7 \times 7} &= \textbf{w}_{3 \times 3}^{71} \cdot \textbf{w}_{3 \times 3}^{72} \cdot \textbf{w}_{3 \times 3}^{73},
\end{aligned}
\end{equation}
where the weight $\textbf{w}_{3 \times 3}^{km}$ correspond to a representation with RF (2\textit{m}+1), we share the representation at the same RF and its corresponding weight, then $\textbf{w}_{7 \times 7} = \textbf{w}_{5 \times 5} \cdot \textbf{w}_{3 \times 3}^{73}$.

In Fig. \ref{representation sharing}, each $p_{i}$ and sphere denote an operation and the decomposed operation, respectively. For example, $p_{2}$ represents Conv $5 \times 5$ with two decomposed operations (two intermediate representations at different RF). We can reuse intermediate representation at the same RF for each factorized operation to share the parameters and computation instead of repetitive computation causing memory problems. Therefore, all decomposed operations reuse a single intermediate representation $r_{3}$ at the RF 3 level, as shown in Fig. \ref{representation sharing}. Notably, the computational numbers of intermediate representations reduce from 22 to 10 after representation sharing, significantly reducing memory usage while speeding up the search process.

\begin{algorithm}[!t]
    \caption{NAS FTI-FDet} \label{algorithm1}
    \KwIn{Data, Searching DAG \begin{math} \mathcal{G} \end{math}, Iteration $T=0$;}
        A mixed operation $o_{i,j}(x)$ created by Eq.~(\ref{equ3}) for each edge $E_{(i,j)}(i < j)$ with same weight $\alpha_{i,j}^{o}$;\\
        Randomly sampling a matrix, equivalenting to a child architecture used for initial training; \\
        \While{not converged and ($T<=20k$)}{
            \eIf{($T\  \% \ 2500==0$)}
            {
                Abandon the lowest potential operation for each edge using Eq.~(\ref{equ5});\\
                Update architecture $\alpha_{i,j}^{o}$ by descending \begin{math} \mathcal{L}_{val}(\theta(\alpha),\alpha) \end{math};
            }
            {
                Train the selected architecture;\\
                Update the weight $\theta$ by descending \begin{math} \mathcal{L}_{train}(\theta,\alpha)\end{math};\\
            }
            $T = T + 1$;
        }
        Decoding final architecture based on the learned $\alpha_{i,j}^{o}$.\\
    \KwOut{Optimal searched architecture. }
\end{algorithm}

\begin{figure}[!t]
	\centering	\includegraphics[width=3.4in]{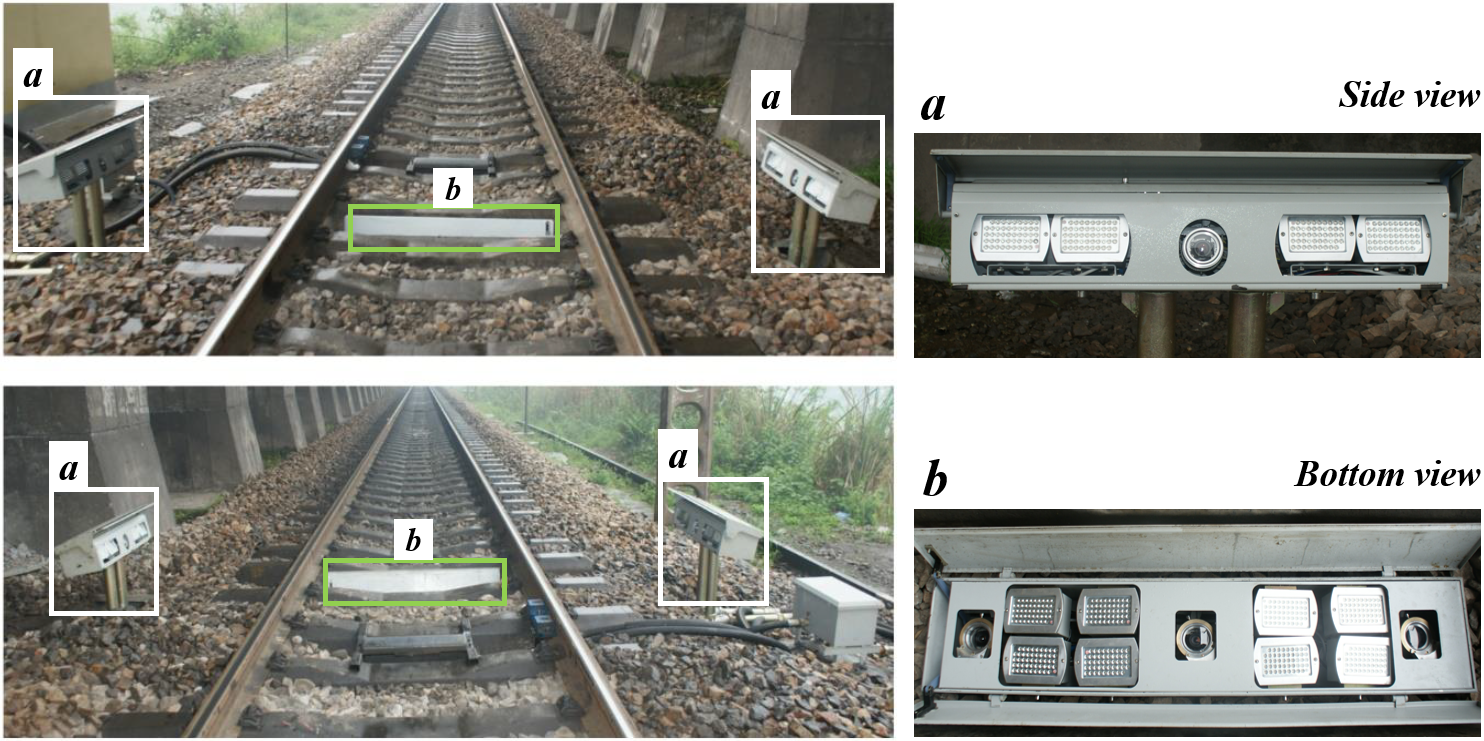}
        \caption{Image acquirement in the wild. (a) Side view. (b) Bottom view. }
	\label{hardware}
\end{figure}

\begin{figure*}[!t]
    \centering
    \includegraphics[width=7.1in]{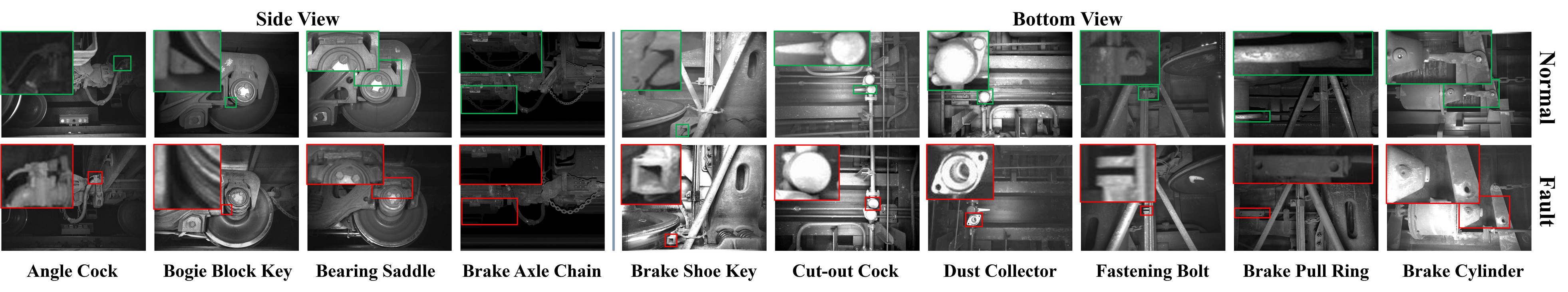}
    \caption{Side View and Bottom View datasets. The top row figures with green ground truth denote normal states for train components, and the bottom row with red ground truth denotes fault states. It is noteworthy that these train components have large-scale variations.} 
	\label{dataset}
\end{figure*}

\subsection{Optimization}
We adopt a continuous relaxation strategy as done in \cite{2018DARTS}, and the mixed operation can be expressed as follows:
\begin{equation}\label{equ5}
    o_{i,j}(x_{i1}) = \sum_{o \in \mathcal{O}} \cfrac{exp(\alpha_{i,j}^{o})}{\sum_{o \in \mathcal{O}}exp(\alpha_{i,j}^{o})}o(x_{i1}),
\end{equation}
where each operation $o_{i,j}$ is assigned with a weight $\alpha_{i,j}^{o}=(\alpha_{1,1}^{o},\alpha_{1,2}^{o},\alpha_{2,1}^{o},\ldots,\alpha_{3,4}^{o})$. The $o$ is all candidates in $\mathcal{O}$. $x_{i1}$ denotes the output of the first $1 \times 1$ convolution within edge. The $o(x_{i1})$ indicates the operation performed on $x_{i1}$. The forward computation of the DAG can be abstracted to matrix multiplication of $9 \times 13$ dimension as follows:
\begin{equation}\label{equ6}
f_{DAG}(x)=
\left[
 \begin{array}{ccc}
     \alpha_{1,1}^{1} & \ldots & \alpha_{1,1}^{13} \\
     \vdots & \ddots & \vdots \\
     \alpha_{3,4}^{1} & \ldots & \alpha_{3,4}^{13} 
 \end{array}
 \right]    
 \otimes
 \left[
  \begin{array}{c}
     Op_1(x) \\
     \vdots \\
     Op_{13}(x) 
 \end{array}
 \right],
\end{equation}
where the left and right matrices represent a sampling matrix and search space, respectively. Each row of the former matrix indicates a mixed operation, and the symbol $\otimes$ defines the interaction process of the two matrices. The operation with the highest probability is retained, while others are abandoned at the derived epoch.
\begin{equation}\label{equ7}
o_{i,j}(x_{i1}) = argmax_{o \in \mathcal{O}}\ \alpha_{i,j}^{o},
\end{equation}

After continuous relaxation, the search process is reduced to a problem that jointly optimizes architecture $\alpha$ and corresponding weights $\theta$. Specifically, the training and validation datasets optimize $\theta$ and $\alpha$, respectively. The bi-level optimization are resolved to this problem as follows:
\begin{equation}\label{equ8}
\begin{aligned}
\underset{\alpha}{min}& \ \mathcal{L}_{val}(\theta(\alpha),\alpha), \\
s.t. \ \theta(\alpha)& = argmin_{\theta} \ \mathcal{L}_{train}(\theta,\alpha).
\end{aligned}
\end{equation}
where \begin{math} \mathcal{L}_{val} \end{math} and \begin{math} \mathcal{L}_{train} \end{math} represent validation and training loss, respectively. Our entire NAS FTI-FDet framework is proposed in the \textbf{Algorithm~\ref{algorithm1}}. 

\begin{table}[!t]
\renewcommand{\arraystretch}{1.2}
	\caption{Fault detection datasets of the freight trains.
		\label{datasets}}
	\centering
		\small
		\setlength{\tabcolsep}{3mm}{
		\begin{tabular}{lccc}
		\toprule
		\multirow{2}{*}{\textbf{Dataset}} &
            \multirow{2}{*}{\textbf{Category}} &
            \multicolumn{2}{c}{\textbf{Total Images}} \\
            \cmidrule{3-4}
            & & train & val \\
            \midrule
            \multirow{4}{*}{\textbf{Side View}} &Angle Cock &1416 &606 \\
            & Brake Axle Chain &244 &105 \\
            & Bogie Block Key  &5440 &2897 \\
            & Load-bearing Saddle  &629 &271 \\
            \midrule
            \midrule
            \multirow{6}{*}{\textbf{Bottom View}} &Brake Cylinder &2723 &1401 \\
            & Brake Pull Ring &1292 &556 \\
            & Brake Shoe Key  &3946 &1690 \\
            & Cut-out Cock &815 &850 \\
            & Dust Collector  &583 &452 \\
            & Fastening Bolt &2050 &1902 \\
                \bottomrule
	\end{tabular}}
\end{table}

\section{Experiments}\label{sec:experiments}
\subsection{Experiments Setup}
\subsubsection{Datasets}
Our fault detection datasets for freight trains contain two scenarios, Side View and Bottom View, which are important for freight trains. All images with a resolution of 700 $\times$ 512 are captured by image acquisition equipment installed along and beside the railway, as shown in Fig. \ref{hardware}. We manually annotate all images with the annotated tool and randomly split 70$\%$ of images as training datasets and 30$\%$ as testing. More details for datasets are shown in Table \ref{datasets} and Fig. \ref{dataset}. 

\subsubsection{Reduced Dataset with Random Sampling}
We divide the datasets into sub-datasets by randomly sampling with five ratios. Let the full dataset: $\mathcal{D}={(s_k,l_k)}$ with $n$ samplings. $s_k$ indicates the $k$-th sample with its corresponding label $l_k$ ($0 \le k \le n$). We take ratio $r \in (1/2,1/4,1/8,1/16,1/32)$ representing the sampling rate on the $\mathcal{D}$. After sampling, the subsets of the dataset are $\mathcal{D}_r \subset \mathcal{D}$ with $n_r$ samplings. The sampling process is
$n_r \approx \sum\nolimits_{(s_k,l_k) \subset \mathcal{D}}^t r \cdot n$,
where $t$ is the number of categories in $\mathcal{D}$, and each sample in its category is selected with equal probability.

\subsubsection{Evaluation Metrics}
To validate the effectiveness of the proposed method, we take average precision (AP) computed over various intersections over union (IoU) thresholds, 
\textit{i.e.}, mAP, AP$_{50}$, AP$_{75}$, as well as AP for small, medium, and large objects are reported to measure the performance of objects with different scales. We also use average recall (AR), which is the maximum recall given a fixed number of detections per image, averaged over categories and IoUs, \textit{i.e.}, AR$_{1}$, AR$_{10}$, AR$_{S}$, AR$_{M}$, and AR$_{L}$. Moreover, we introduce some indexes to verify the suitability for our task, \textit{i.e.}, search cost (GPU days), testing memory usage, and model size. Among them, the search cost indicates search efficiency. Memory usage during testing reflects the dependence of detectors on hardware. The model size measures calculation cost. Our task lies in a scenario that is constrained by limited hardware resources. Therefore, the latter three metrics are also essential.

\subsubsection{Implementation Details}
We conduct experiments on the MS COCO~\cite{2014Microsoft} and our fault datasets. The size of the input images remained at the original size of 700$\times$512. For MS COCO, we take 1$\times$ training strategy using 2 NVIDIA GTX3090 GPU with a batch size of 24 images per GPU. The initial learning rate is 0.02, and we use an SGD optimizer with 0.9 momentum and 0.0001 weight decay. For our fault datasets, we conduct all experiments for searching and training using a single NVIDIA GTX3090 GPU. During the search process, we take the setting of cell $L=1$, nodes $N=3$, and channels $C=96$. The searching process is set to 10 epochs with a batch size of 6 on all datasets. When searching on the whole bottom view dataset, we set 20K max iterations and derive architecture every 2.5K iterations where the max iterations are calculated by the number of training images and required divisible by derived iteration that is adjusted by the dataset size. We use the Adam optimizer with a learning rate of 0.004 and weight decay of 0.0001. The learning rate reduces to 1/10 at the 3/4 max iterations. During the training process, we take a 2$\times$ training schedule for lightweight backbones and 1$\times$ for others. Moreover, we only change $L = 2$ compared with the searching process, and other training configurations are the same as those in FCOS~\cite{FCOS}. It is worth noting that all models searched on different sub-datasets are trained on the whole dataset.


\begin{table}[!t]
	\renewcommand{\arraystretch}{1.2}
	\caption{Comparisons of different micro-structures in the searched module. The results are obtained with RFDNet on the Bottom View dataset.}
	\centering
	\label{searchable module}
	\small
	\setlength{\tabcolsep}{1.5mm}{
	\begin{tabular}{cccccccc}
		\toprule
		\multirow{2}{*}{\textbf{Cells}} & 
		\multirow{2}{*}{\textbf{Nodes}} & 
            \multirow{2}{*}{\textbf{mAP}} &   
            \multirow{2}{*}{\textbf{AP$_{50}$}} &
            \multirow{2}{*}{\textbf{AP$_{75}$}} &
            \multirow{2}{*}{\textbf{AR$_{1}$}} &
            \multirow{2}{*}{\textbf{AR$_{10}$}} &
            \multirow{2}{*}{\textbf{\makecell[c]{Model\\Size(MB)}}}\\
            & & & & & & \\
		\midrule
		\multirow{3}{*}{\textbf{1}} &2 &43.0 &68.4 &47.1&59.3  &60.8  &35.6  \\
		   &3 &44.3 &71.3 &49.6 &60.2 &61.5 &38.5   \\
              &4 &44.7 &70.3 &50.4 &60.7  &62.1  &39.3  \\
		\midrule
		\multirow{3}{*}{\textbf{2}} &2 &44.0 &69.7 &49.1 &60.1 &61.2  &41.4  \\
		   &3 &45.0 &71.0 &51.4 &61.2 &62.6 &48.3   \\
              &4 &44.6 &70.8 &50.1 &61.2 &62.5 &48.9 \\ 
		\midrule
		\multirow{3}{*}{\textbf{3}} &2 &43.0 &68.9 &48.5 &59.9 &61.4 &48.3  \\
		   &3 &42.6 &67.6 &47.9 &60.6 &62.0 &57.6  \\
              &4 &42.1 &67.7 &47.3 &60.2 &61.3 &60.5 \\
		\bottomrule 
	\end{tabular}}
\end{table}

\subsection{Ablation Study}\label{sec:ablation}

\begin{figure*}[!t]
	\centering
	\subfigure[Metric mAP]{
       \includegraphics[width=1.7in]{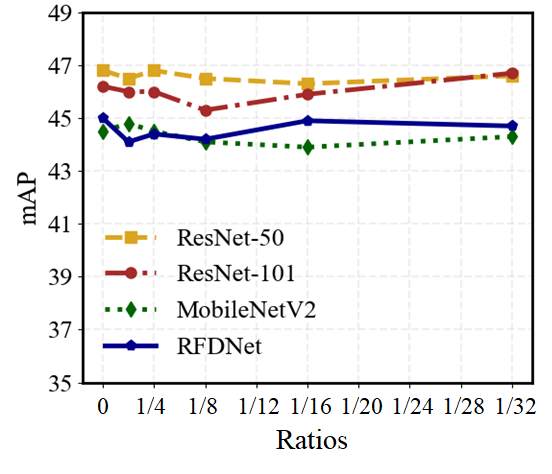}
	}\hspace{-0.6em}
    \subfigure[Metric AP$_{50}$]{
        \includegraphics[width=1.7in]{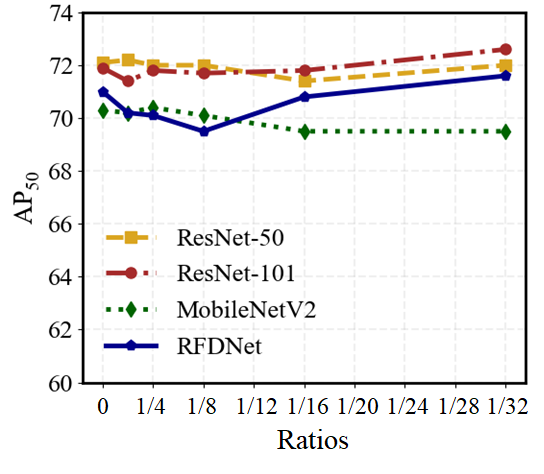}
	}\hspace{-0.6em}
	\subfigure[Metric AR$_1$]{
       \includegraphics[width=1.7in]{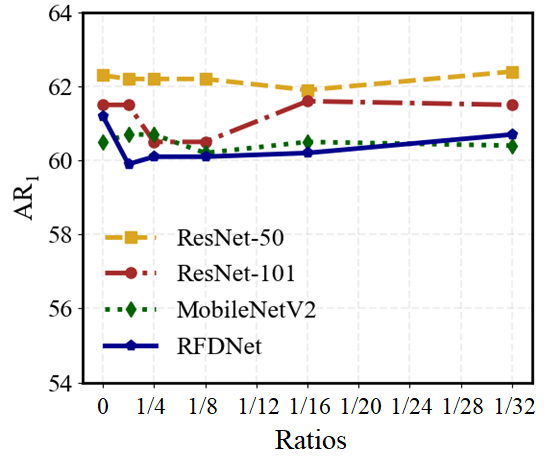}
	}\hspace{-0.6em}
    \subfigure[Search Cost]{
        \includegraphics[width=1.7in]{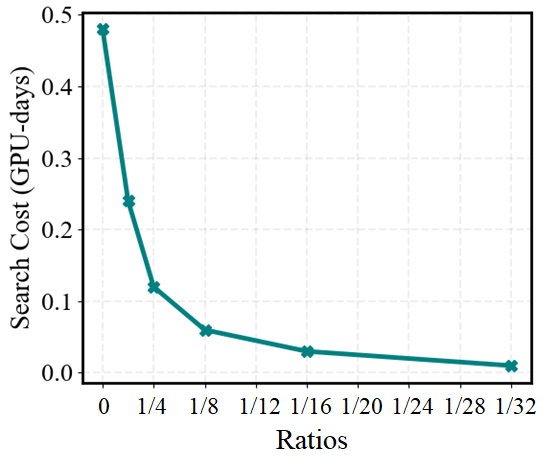}
	}\vspace{-0.6em}
	\caption{Performance of models searched on Bottom View dataset with different ratios under various backbones. Note that the search cost is related to data volumes and search space but not backbones. Our framework can keep consistent performance under different data volumes, namely having data volume robustness.
	}
        \label{Data Volume Robustness}
\end{figure*}

\subsubsection{Micro-Structure of Searched Module}
The micro-structure of the searched module consists of two hyper-parameters: the number of nodes $N$ and the repeated cells $L$. Each node represents a layer, and its number defines the depth of cells, which decides the depth of the searched network in turn. The two hierarchical variables jointly influence the searched structure. By varying $N$ and $L$, our objective is to enhance comprehension of how the micro-structure influences the quality of the architectures that we uncover. The performance of different combinations of nodes and cells is shown in Table \ref{searchable module}. Quantity does not mean quality, such as $L = 3$. More cells, namely a more deep network, cause more information loss. With the same number of cells, precision metrics sometimes vary greatly due to the better operation it chose. Owing to DS convolution, the model size has a slight difference with increasing the number of nodes at times. The model with $L = 2$ and $N = 3$ reaches a better compromise between performance and model size. Thus, we take this setting during all experiments in this work.

\begin{table}[!t]
	\renewcommand{\arraystretch}{1}
	\caption{Ablation experiments under different search spaces with ResNet-50 backbone on the Bottom View dataset.}
	\label{Search Space}
 	\centering
	\setlength{\tabcolsep}{2pt}{
		\small
		\setlength{\tabcolsep}{1mm}{
		\begin{tabular}{ccccccc}
			\toprule
		  \multirow{1}{*}{\textbf{Method}} &
            \multirow{1}{*}{\textbf{\makecell[c]{Search Space}}}& 
		\multirow{1}{*}{\textbf{mAP}}&  
            \multirow{1}{*}{\textbf{AP$_{50}$}}&
            \multirow{1}{*}{\textbf{AP$_{75}$}}&
		\multirow{1}{*}{\textbf{AR$_1$}}&
            \multirow{1}{*}{\textbf{AR$_{10}$}}
            \\
			\midrule
		  Baseline &- &46.1 &71.3 &52.3 &61.9 &62.6  \\
            NAS FTI-FDet &\begin{math}  \mathcal{O} \end{math}  &46.8 &72.0 &53.2 &62.6 &63.4  \\
            NAS FTI-FDet &\ \ \begin{math}  \mathcal{O}_{s1} \end{math}  &46.7 &72.6 &53.0 &61.7 &62.8  \\
            NAS FTI-FDet &\ \ \begin{math} \mathcal{O}_{s2} \end{math} &46.6 &71.7 &52.8 &62.3 &63.0   \\
			\bottomrule
	\end{tabular}}}
\end{table}

\begin{table}[!t]
	\renewcommand{\arraystretch}{1}
	\caption{Effectiveness of search space for scale variation on the Side View dataset.}
	\label{scale}
 	\centering
	\setlength{\tabcolsep}{2pt}{
		\small
		\setlength{\tabcolsep}{0.5mm}{
		\begin{tabular}{ccccccccc}
			\toprule
            \multirow{2}{*}{\textbf{Method}}& 
		\multirow{2}{*}{\textbf{\makecell[c]{Search\\Space}}}&  
            \multirow{2}{*}{\textbf{mAP}}&
            \multirow{2}{*}{\textbf{AP$_S$}}&
		\multirow{2}{*}{\textbf{AP$_M$}}&
            \multirow{2}{*}{\textbf{AP$_L$}}&
            \multirow{2}{*}{\textbf{AR$_S$}}&
            \multirow{2}{*}{\textbf{AR$_M$}}&
            \multirow{2}{*}{\textbf{AR$_L$}}
            \\
            &&&&&&&&\\
			\midrule
            Baseline  &- &47.3 &30.1 &56.1 &35.2 &30.0 &64.9 &51.5  \\
            NAS FTI-FDet &\begin{math}\mathcal{O} \end{math} &47.9 &30.2 &58.5 &39.3 &30.0 &67.5 &54.0  \\
			\bottomrule
	\end{tabular}}}
\end{table}

\subsubsection{Effectiveness of Search Space}
Different tasks are sensitive to different search spaces. Some experiments are designed to verify the impact of the proposed search spaces. We take FCOS as the baseline and search for the head to replace the original one for comparison. In Table \ref{Search Space}, our search space boosts the performance by 0.7 mAP compared to the baseline while improving all other indexes. As mentioned, the FCOS remains the same structure for regression and classification, while the searched cells for the two subnets are distinct, which may bring benefits due to the different information required for the two tasks. To further study the influence of RF, we search on the two sub-search spaces: $\mathcal{O}_{s1}$ and $\mathcal{O}_{s2}$, indicating the search space with RF smaller than 5 and 7, respectively. We observe that the accuracy performance is lower on $\mathcal{O}_{s1}$ than $\mathcal{O}_{s2}$ but at opposite in recall, and the performance on the two sub-space is not as good as the one on whole $\mathcal{O}$. These results suggest that appropriate RF in layers is crucial for fault detection. Moreover, we present results on the Side View dataset in Table \ref{scale}, which indicates that our search space can remedy the large characteristic difference (scale variation) between inter-classes.

\begin{table}[!t]
	\renewcommand{\arraystretch}{1}
	\caption{Comparison of different backbones on Bottom View dataset. MNetV2 is MobileNetV2~\cite{2018MobileNetV2}, and Base denotes the baseline.
		\label{backbone}}
	\centering
	\small
	\setlength{\tabcolsep}{1mm}{
		\begin{tabular}{lccccccc}
			\toprule
                \multirow{2}{*}{\textbf{Backbone}}& \multirow{2}{*}
            {\textbf{Method}}&
			\multirow{2}{*}
            {\textbf{mAP}}& 
			\multirow{2}{*}
            {\textbf{AP$_{50}$}}& 
			\multirow{2}{*}
            {\textbf{AP$_{75}$}}&
			\multirow{2}{*}
            {\textbf{AR$_{1}$}}&
			\multirow{2}{*}
            {\textbf{AR$_{10}$}}& 
                \multirow{2}{*}
            {\textbf{\makecell[c]{Model\\Size(MB)}}}\\
            & & & & & & & \\
			\midrule
            \multirow{2}{*}{ResNet-50}
            &Base &45.9 &70.2 &52.2 &60.9 &61.7 &128.9 \\
              &Ours &46.8 &72.0 &53.2 &62.6 &63.4 &126.6 \\
            \multirow{2}{*}{ResNet-101}
            &Base &45.6 &69.6 &50.6 &60.8 &61.4 &205.2 \\
              &Ours &46.2 &71.9 &52.0 &61.5 &63.1 &202.1 \\
            \multirow{2}{*}{MNetV2}
            &Base &43.6 &67.2 &50.2 &59.8 &60.8 &38.8 \\
              &Ours &45.2 &71.0 &50.7 &61.2 &62.4 &42.6 \\
            \multirow{2}{*}{RFDNet}
            &Base &43.1 &67.1 &48.6 &58.4 &59.3 &48.7 \\
              &Ours &45.0 &71.0 &51.4 &61.2 &62.6 &48.3 \\
			\bottomrule
	\end{tabular}}
\end{table}

\subsubsection{Generality on Different Backbone}
We use different backbones including lightweight ones such as MobileNetV2~\cite{2018MobileNetV2}, RFDNet~\cite{RFDNet}, and complex ones like ResNet-50/101~\cite{resnet-50, resnet-101}, to verify the generality of our proposed framework. As shown in Table~\ref{backbone}, we all get improvements under different backbones compared with the baseline (FCOS). This suggests that our method can search for better networks and have generality for the backbones with diverse capacities. Our model has less model size, which lies in the channel size settings (96 in our sub-networks and 256 in FCOS). This confirms the fact that large channel sizes may cause feature redundancy. It is also possible that more information interactions between nodes (in Fig. \ref{framework}) compensate for the decrease in channel size. Among them, ResNet-50 has the best performance in all metrics except model size. In terms of lightweight backbones, RFDNet achieves 45.0 mAP and 61.2 AR$_1$ compared to 44.5 mAP and 60.5 AR$_1$ by MobileNetV2, and the former model size is 12.9MB larger than the latter. It indicates that lightweight backbones achieve a much better tradeoff between resources and accuracy, making them more suitable for resource-constrained environments.

\begin{figure*}[!t]
	\centering
	\subfigure[Original]{
       \label{o}
       \includegraphics[width=1.7in]{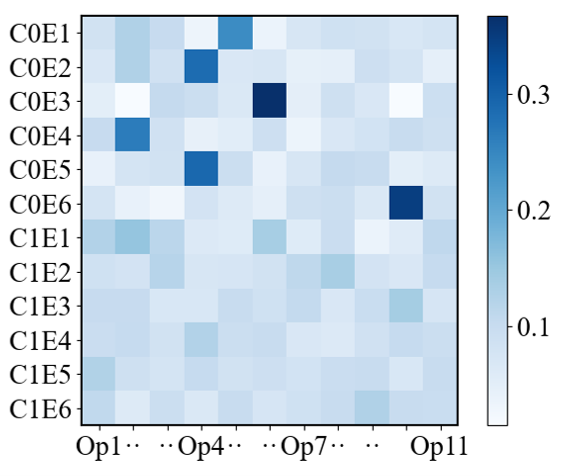}
	}\hspace{-0.6em}
    \subfigure[Ratio 1/2]{
        \label{1/2}
        \includegraphics[width=1.7in]{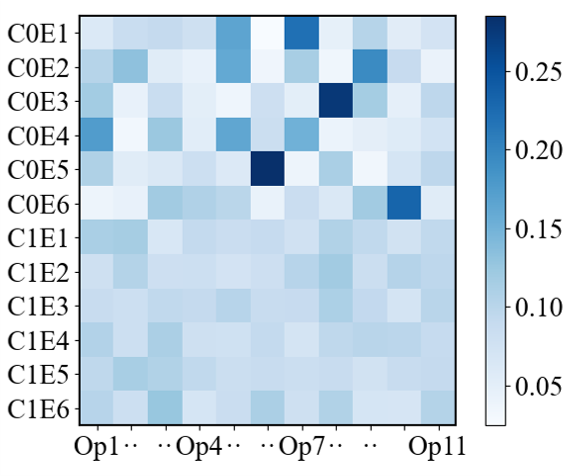}
	}\hspace{-0.6em}
	\subfigure[Ratio 1/4]{
       \label{1/4}
       \includegraphics[width=1.7in]{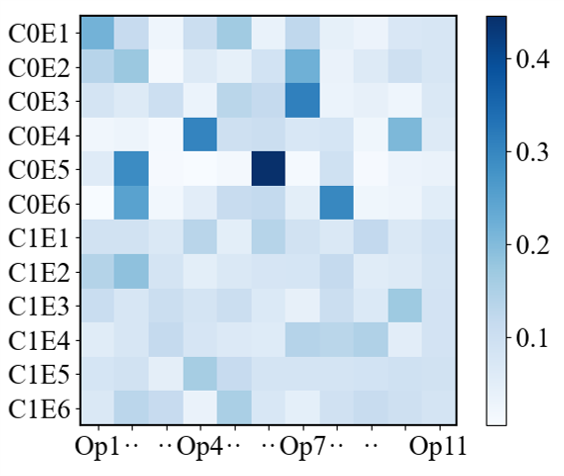}
	}\hspace{-0.6em}
    \subfigure[Cell Legend]{
        \label{legend}
        \includegraphics[width=1.7in]{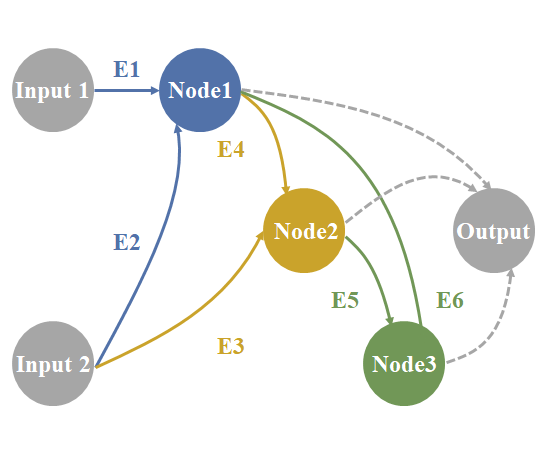}
	}\vspace{-0.6em}
	\caption{The probability heatmap on the two reduced Bottom View datasets, as shown in \subref{o}-\subref{1/4}. The color bar represents the probability. Each row corresponds to the edge within the specific cell, and each column corresponds to operations $Op_k$ in $\mathcal{O}$. The C0 and C1 denote the cell for regression and classification, respectively. Each intermediate node has two edges, as shown in \subref{legend}.
	}
        \label{Operation Decisions}
\end{figure*}

\subsubsection{Time Savings and Data Volume Robustness}
The results of the performance of the searched network on five reduced datasets and search cost are illustrated in Fig. \ref{Data Volume Robustness}. For the metric mAP, the trend of the folding line of each backbone generally remains stable. In terms of AR$_1$, the folding lines relating to ResNet-101 slightly fluctuate at the Ratios 1/4 and 1/8. The RFDNet on the whole dataset has the highest AR$_1$ and maintains stable performance on other sub-datasets. We find that some searched architectures on reduced datasets perform better than those searched on the whole dataset. This is probably because the reduced data can satisfy the recognition of intra-class components with similar characteristics, or some samplings are negative to the search process. On the other hand, perhaps our proposed search space has enough robustness, which makes up for the impact of reduced samplings. We can observe a linear dependency between search cost and dataset size. The above analysis reveals that our method is robust for different data volumes and the effectiveness of decreasing search time from the data volume direction.

\subsubsection{Operation Decisions under Different Data Volumes}
We are interested in exploring how operation choices with different RF change when applied to sub-datasets with different data volumes. Therefore, we reduce the None operation here, and $Op_{11}$ denotes the Def Conv in Fig. \ref{Operation Decisions}. We derive the final probability heatmap of operations selected for each edge and only choose two reduced datasets for simplicity. The edge distribution of two cells for classification and regression is shown in Fig. \ref{Operation Decisions}.
It has been found that the probability of operations within the cell for classification (C1) has little difference, while the one for regression (C0) at opposite. The reason for this phenomenon probably lies in regression being sensitive to these operations with different RF, while classification on the contrary. Furthermore, it is probably because of our connection type of sub-networks, in which the first cell is for regression and the cascade connection between the first and second cell is for classification. The regression cell has chosen appropriate RF targeting at the scale distribution of train components and provided enough comprehensive features, consequently affecting the subsequent decisions by classification. Besides, final operation decisions vary with data volumes because of different data characteristics.

\begin{table}[!t]
	\renewcommand{\arraystretch}{1.2}
	\caption{Comparison of different methods including hand-crafted and NAS-based ones on the MS-COCO dataset. The backbone of all the models is ResNet-50.}
		\label{generalzation}
	\centering
        \setlength{\tabcolsep}{2pt}{
	\small
	\setlength{\tabcolsep}{1mm}{
		\begin{tabular}{lcccccc}
			\Xhline{1pt}
            \textbf{Methods}& 
            \textbf{mAP}&
            \textbf{AP$_{50}$}&
            \textbf{AP$_{75}$}&
            \textbf{AP$_{S}$}&
            \textbf{AP$_{M}$}&
            \textbf{AP$_{L}$}\\
                \Xhline{0.5pt}
        FCOS	\cite{FCOS}	  
            &34.3 &52.4 &36.6 &16.3 &38.1
            &49.5
            \\
        Faster R-CNN \cite{Faster_R-CNN}		 
            &34.1 &54.8 &36.1 &16.0 &37.8 
            &48.5 
            \\
        NAS-FCOS \cite{NAS-FCOS}   
            &33.3 &51.7 &35.2 &14.6 &37.0
            &47.8 
            \\
        NAS FTI-FDet 
            &34.8 &52.8 &37.3 &16.4 &38.7
            &49.8 
            \\ \Xhline{1pt}
	\end{tabular}}}
\end{table}

\subsubsection{Generalization on Other Datasets}
To evaluate the generalization of our method on other datasets, we compare the model searched on our fault dataset with hand-crafted and NAS-based methods on the MS-COCO dataset in Table~\ref{generalzation}. All methods adopt ResNet-50 as the backbone. Our NAS FTI-FDet surpasses baseline FCOS by 0.5 mAP, indicating the transferability of our architecture on different datasets. It is probably due to the MS COCO being similar to our fault detection dataset with large-scale variations, and our scale-aware search space is designed to remedy this problem. Moreover, the mAP improvement for small, medium, and large objects are 0.1, 0.6, and 0.3 compared with FCOS, respectively. The performance gain of large objects is better than that of small objects, possibly due to the selected operation with larger RF. Compared with the two-stage detector Faster R-CNN with RPN, the proposed NAS FTI-FDet also has advantages that indicate the effectiveness of discovered sub-networks with multi-scale representation. And our method dominates the NAS-based method NAS-FCOS as well. These results further demonstrate the generalization of our method on other datasets.

\begin{table*}[!t]
	\renewcommand{\arraystretch}{1.3}
	\caption{Comparison with The SOTA Methods on The Whole Typical Fault Datasets.
		\label{sota}}
	\centering
	\small
        \begin{threeparttable}
	\setlength{\tabcolsep}{1.2mm}{
		\begin{tabular}{l|c|ccccc|ccccc|c|c}
			\Xhline{1pt}
			\multirow{3.2}{*}{\textbf{Methods}}& 
			\multirow{3.2}{*}{\textbf{Backbone}}&
                \multicolumn{10}{c|}{Datasets}&
                \multirow{3.2}{*}{\makecell[c]{Model size\\(MB)}} &
                \multirow{3.2}{*}{\makecell[c]{Test Memory$^{*}$\\(MB)}}
                \\
                \cline{3-12}
                & &
                \multicolumn{5}{c|}{Bottom View}&
                \multicolumn{5}{c|}{Side View}&
                &  \\
                \cline{3-12}
                & & mAP &AP$_{50}$ &AP$_{75}$ &AR$_{1}$ & AR$_{10}$&
                    mAP &AP$_{50}$ &AP$_{75}$ &AR$_{1}$ & AR$_{10}$&
                &  \\
                \Xhline{0.5pt}
        CenterNet \cite{Centernet} &ResNet-18  
            &43.8 &70.6 &47.2 &58.8 &59.5 
            &40.7 &67.0 &40.8 &52.0 &54.4 
            &66.8 
            &1565
            \\ 
        CentripetalNet \cite{CentripetalNet} &HNet-104 
            &38.7 &63.6 &42.1 &60.5 &60.5 
            &38.8 &63.0 &39.1 &51.7 &53.1 
            &832.8 
            &2835
            \\
        FCOS	\cite{FCOS}	 &ResNet-50  
            &45.9 &70.2 &52.2 &60.9 &61.7
            &\textbf{47.3} &\textbf{73.8} &50.1 &55.7 &58.2 
            &128.9 
            &1665 
            \\
        GFL \cite{GFL}        &ResNet-50
            &45.9 &\textbf{71.4} &52.1 &61.7 &62.9
            &46.9 &72.7 &48.8 &55.6 &59.0 
            &130.0 
            &1705 
            \\
        RetinaNet \cite{Retinanet}  &ResNet-18 
            &35.8 &62.5 &37.2 &50.3 &57.0
            &37.0 &63.8 &37.3 &49.9 &53.6
            &81.5 
            &1647
            \\
        RetinaNet \cite{Retinanet} &SwinT 
            &42.5 &67.1 &48.4 &60.3 &63.7
            &40.1 &67.2 &42.4 &52.5 &55.9 
            &151.0 
            &1823  
            \\
        YOLOF \cite{YOLOF}   &ResNet-50 
            &45.8 &71.0 &52.6 &59.3 &62.2
            &47.2 &73.8 &\textbf{50.9} &55.1 &58.6 
            &171.7 
            &1701 
            \\
        YOLOX
            \cite{YOLOX}   &CSPNet
            &45.9 &70.0 &51.4 &60.2 &61.9
            &44.4 &69.3 &46.3 &55.5 &57.3 
            &84.9 
            &1649 
            \\
        YOLOv7
            \cite{wang2023yolov7}   &CSPNet
            &36.4 &57.4 &41.4 &54.3 &57.7
            &44.6 &69.4 &47.0 &\textbf{55.9} &\textbf{59.4}
            &12.3 
            &1399 
            \\
        YOLOv8
            \cite{yolov8}   &CSPNet
            &42.6 &65.0 &48.1 &56.8 &57.8
            &44.2 &67.8 &46.7 &55.4 &59.2
            &\textbf{6.3} 
            &\textbf{1385}
            \\
        Faster R-CNN \cite{Faster_R-CNN}		&ResNet-50 
            &\textbf{46.0} &70.5 &\textbf{52.8} &60.4 &62.5 
            &41.0 &66.1 &42.9 &52.3 &53.3 
            &167.7 
            &1935 
            \\
        Libra R-CNN 
            \cite{Libra_R-cnn}	&ResNet-18  
            &40.6 &67.0 & 44.7&56.9 &62.0 
            &39.9 &66.3 &42.3 &52.8 &55.6 
            &82.6 
            &1889 
            \\
        Sparse R-CNN	\cite{Sparse_R-CNN}	&ResNet-50 
            &42.7 &68.0 &47.9 &56.9 &60.0
            &41.0 &67.1 &42.2 &52.2 &53.1 
            &269.0 
            &1829
            \\
        Deform. DETR \cite{Deformable_DETR}	&ResNet-50 
            &36.1 &68.3 &34.1 &\textbf{62.2} &\textbf{65.0} 
            &34.1 &65.3 &32.4 &47.2 &51.0
            &210.7 
            &1727 
            \\ \hline
        NAS-FPN \cite{NAS-FPN}  &ResNet-50
            &42.4 &69.3 &46.5 &55.5 &59.2 
            &46.7 &72.5 &48.9 &54.0 &56.9
            &314.4 
            &3767 
            \\
        NAS-FCOS \cite{NAS-FCOS}   &ResNet-50
            &45.4 &69.6 &51.6 &60.0 &61.3
            &47.5 &72.8 &50.6 &55.4 &59.1 
            &157.6 
            &1751 
            \\
        DetNAS \cite{DetNAS}   &Searched
            &42.8 &67.4 &48.6 &57.2 &59.5 
            &45.9 &71.9 &49.0 &53.8 &55.4
            &162.1 
            &2892 
            \\
        OPANAS \cite{2021OPANAS}   &ResNet-50
            &45.2 &70.2 &52.4 &59.4 &60.4
            &44.6 &70.6 &47.1 &54.7 &55.7
            &144.0 
            &1394 
            \\
        FAD$^\ddag$ \cite{FAD}              &MNetV2 
            &44.5 &70.3 &50.9 &60.5 &61.8 
            &43.4 &68.8 &46.3 &55.2 &59.8 
            &68.6 
            &\textbf{1251}
            \\
        NAS FTI-FDet$^\ddag$ (ours)  &RFDNet 
            &45.0 &71.0 &51.4 &61.2 &62.6 
            &42.0 &66.7 &45.0 &54.6 &59.1
            &48.3 
            &1283 
            \\
        NAS FTI-FDet$^\ddag$ (ours) &MNetV2
            &45.2 &71.0 &50.7 &61.2 &62.4 
            &44.9 &70.4 &\textbf{54.6} &54.6 &59.0
            &\textbf{42.6} 
            &1267 
             \\
        NAS FTI-FDet \;\! (ours)  &ResNet-50 
            &\textbf{46.8} &\textbf{72.0} 
            &\textbf{53.2} &\textbf{62.6} &\textbf{63.4} 
            &\textbf{47.9} &\textbf{72.9} & 51.5 &\textbf{55.7} &\textbf{59.9}
            &126.6 
            &1375
            \\ \Xhline{1pt}
	\end{tabular}}
        \begin{tablenotes}
            \footnotesize
        \item[ ] MNetV2, SwinT, CSPNet, and HNet-104 represent MobileNetV2~\cite{2018MobileNetV2}, SwinTransformer \cite{swin}, CSPDarkNet \cite{Cspnet}, and Hourglass-104 \cite{CornerNet}, respectively.  
        \item[*] The test memory is evaluated on a single NVIDIA GTX2080Ti GPU using 1 image per batch.
        \item[\ddag] means the 2x schedule in training.
      \end{tablenotes}
    \end{threeparttable}
\end{table*}

\subsection{Comparison with SOTA Methods}
To further demonstrate the advantages of our framework, we make comparisons with SOTA methods, including some NAS-based and hand-crafted ones. The experimental results are summarized in Table \ref{sota}.

\subsubsection{Comparison with Hand-Crafted Methods}
On the Bottom View dataset, our method with ResNet-50 achieves 46.8 mAP, outperforming almost all hand-crafted methods with the same counterpart. It also dominates on other metrics except for AR$_{10}$. Interestingly, Deform. DETR achieves competitive performance on recall, but perform unsatisfactory on the accuracy. On the Side View dataset, our method with ResNet-50 also surpasses all detectors (47.9 mAP). In addition, our method achieves 0.9 and 0.6 mAP higher than FCOS on the Bottom and Side View datasets, respectively, with less than 17.4 \% test memory and nearly equal model size. These results suggest that our method can automatically design better networks than hand-crafted ones.

\subsubsection{Comparison with NAS-Based Methods}
Our proposed NAS FTI-FDet with ResNet-50 obtains 46.8 and 47.9 mAP on Bottom View and Side View datasets, respectively, which outperforms all NAS-based models. Moreover, our method on the two datasets concerning accuracy and recall also obtain competitive results. As for the model size and test memory, our method with ResNet-50 obtains at least 12.1\% less model size than others and consumes the smallest test memory. For lightweight backbones, the model size of our proposed model with MobileNetV2 is 26 MB lower than FAD and a bit of test memory higher than FAD. It is noteworthy that the accuracy has some sacrifices by replacing with lightweight backbones, such as MobileNetV2 and RFDNet, yet a nearly 60$\%$ reduction in model size. For example, our methods with MobileNetV2 and ResNet-50 have 42.6 MB and 126.6 MB model sizes, respectively. Regarding search cost, our method is more efficient than others, as shown in Table \ref{search method}. The results show that our method is much more efficient and effective.

\begin{figure}[!t]
    \centering	
    \includegraphics[width=3.4in]{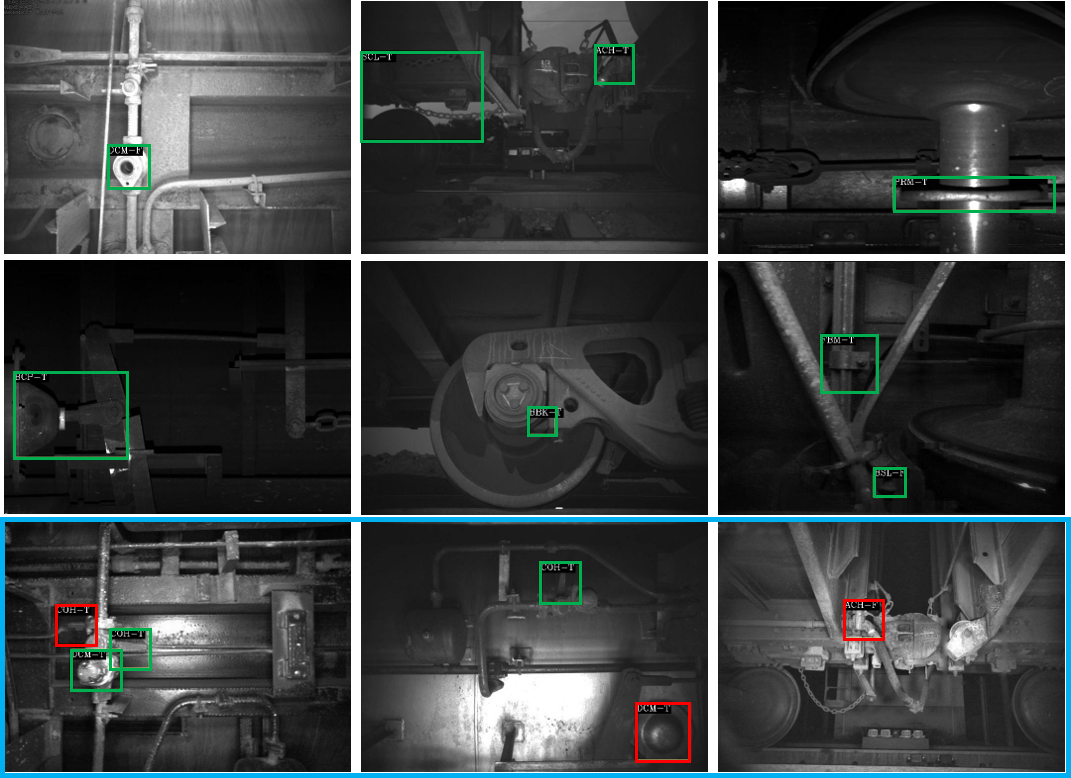}
    \caption{Visual representations of the experimental results. The green and red box represent correct detection and false detection, respectively. The robustness of our proposed method is not satisfactory in some complex situations.}
    \label{false detection}
\end{figure}

\subsection{Discussion}
During the process of exploring data volume robustness, the operation decision varies with dataset size because different data distributions result in different RF combinations. We can adjust the search space based on the preference of operation selection for adapting the data characteristic to guarantee data volume robustness. Moreover, the reduced dataset size can also bring gains to the NAS. The searched models oftentimes perform better on sub-datasets instead of on the whole dataset, and the search cost is linearly decreased by the dataset size. It is a significant observation to ensure the performance and fast iteration of the models using decreased datasets when capturing vast amounts of new data in fault detection. In addition, the visual results are shown in Fig. \ref{false detection}, and our method has some false detection under uneven illumination conditions. To enhance the robustness, a straightforward method is used to expand these types of images or adopt data augmentation. From the perspective of NAS, we can design an attention-based search space to adaptively focus on the fault components under uneven illumination conditions. Consequently, we will work towards these ideas to ensure robustness in the face of these challenges.

\section{Conclusion}\label{sec:conclusion} 
In this paper, we propose the NAS FTI-FDet framework to efficiently search for an optimal detection head with multi-scale representation.
In particular, we introduce a scale-aware search space with more RF candidate operations to explore the effective RF for fitting data characteristics. An efficient sharing method is further proposed to reduce memory limitation caused by the large search space and improve search efficiency.
Experiments demonstrate that our method achieves competitive performance, 46.8 mAP/62.6 AR$_1$ on the Bottom View dataset and 47.9 mAP/55.7 AR$_1$ on the Side View dataset, respectively. Moreover, our proposed method shows data volume robustness that maintains comparable or even better accuracy with reduced data volumes, linearly decreasing the search cost.

In our future work, we intend to use different sampling methods to analyze the role of samplings in the search process. We plan to improve illumination robustness and detection speed while maintaining accuracy. Moreover, the backbone is also critical in object detection. We are working on the optimization of the detection backbone with a highly efficient training-free NAS-based method by a multi-objective evolutionary algorithm, targeting the expressivity, trainability, and complexity of the neural network.

\small
\bibliographystyle{ieeetr}
\bibliography{Bibliography/IEEEabrv,Bibliography/BIB_xx-TII-xxxx}

\begin{thebibliography}{10}

\bibitem{butera2021precise}
L.~Butera, A.~Ferrante, M.~Jermini, M.~Prevostini, and C.~Alippi, ``Precise
  agriculture: Effective deep learning strategies to detect pest insects,''
  {\em IEEE-CAA J. Automatica Sin.}, vol.~9, no.~2, pp.~246--258, 2021.

\bibitem{zhang2022visual}
Y.~Zhang, Y.~Zhou, H.~Pan, B.~Wu, and G.~Sun, ``Visual fault detection of
  multi-scale key components in freight trains,'' {\em IEEE Trans. Ind.
  Informat.}, vol.~19, pp.~9082--9090, 2022.

\bibitem{saufi2020gearbox}
S.~R. Saufi, Z.~A.~B. Ahmad, M.~S. Leong, and M.~H. Lim, ``Gearbox fault
  diagnosis using a deep learning model with limited data sample,'' {\em IEEE
  Trans. Ind. Informat.}, vol.~16, no.~10, pp.~6263--6271, 2020.

\bibitem{2.A.2}
J.~Sun, Y.~Xie, and X.~Cheng, ``A fast bolt-loosening detection method of
  running train's key components based on binocular vision,'' {\em IEEE
  Access}, vol.~7, pp.~32227--32239, 2019.

\bibitem{2.A.4}
C.~Chen, X.~Zou, Z.~Zeng, Z.~Cheng, L.~Zhang, and S.~C.~H. Hoi, ``Exploring
  structural knowledge for automated visual inspection of moving trains,'' {\em
  IEEE Trans. Cybern.}, vol.~52, no.~2, pp.~1233--1246, 2022.

\bibitem{2.A.5}
Z.~Zhou, Y.~Hu, X.~Deng, D.~Huang, and Y.~Lin, ``Fault detection of train
  height valve based on nanodet-resnet101,'' in {\em YAC}, pp.~709--714, 2021.

\bibitem{NAS-FPN}
G.~Ghiasi, T.~Lin, and Q.~V. Le, ``Nas-fpn: Learning scalable feature pyramid
  architecture for object detection,'' in {\em CVPR}, pp.~7036--7045, 2019.

\bibitem{NAS-FCOS}
N.~Wang, Y.~Gao, H.~Chen, P.~Wang, Z.~Tian, C.~Shen, and Y.~Zhang, ``Nas-fcos:
  Fast neural architecture search for object detection,'' in {\em CVPR}, 2020.

\bibitem{DetNAS}
Y.~Chen, T.~Yang, X.~Zhang, G.~Meng, X.~Xiao, and J.~Sun, ``Detnas: Backbone
  search for object detection,'' in {\em NeurIPS}, pp.~6638--6648, 2019.

\bibitem{2021OPANAS}
T.~Liang, Y.~Wang, Z.~Tang, G.~Hu, and H.~Ling, ``{OPANAS:} one-shot path
  aggregation network architecture search for object detection,'' in {\em
  CVPR}, pp.~10195--10203, 2021.

\bibitem{2.A.3}
Y.~Sun, N.~Qin, and L.~Ma, ``High-speed train bogie faults diagnosis using
  singular spectrum analysis,'' in {\em DDCLS}, pp.~56--59, 2018.

\bibitem{Centernet}
X.~{Zhou}, D.~{Wang}, and P.~{Kr{\"a}henb{\"u}hl}, ``{Objects as Points},''
  {\em arXiv preprint arXiv:1904.07850}, 2019.

\bibitem{CentripetalNet}
Z.~Dong, G.~Li, Y.~Liao, F.~Wang, P.~Ren, and C.~Qian, ``Centripetalnet:
  Pursuing high-quality keypoint pairs for object detection,'' in {\em CVPR},
  pp.~10519--10528, 2020.

\bibitem{FCOS}
Z.~Tian, C.~Shen, H.~Chen, and T.~He, ``{FCOS: Fully Convolutional One-Stage
  Object Detection},'' in {\em ICCV}, pp.~9626--9635, 2019.

\bibitem{GFL}
X.~Li, W.~Wang, L.~Wu, S.~Chen, X.~Hu, J.~Li, J.~Tang, and J.~Yang,
  ``Generalized focal loss: Learning qualified and distributed bounding boxes
  for dense object detection,'' {\em NeurIPS}, vol.~33, pp.~21002--21012, 2020.

\bibitem{Retinanet}
T.-Y. Lin, P.~Goyal, R.~Girshick, K.~He, and P.~Doll{\'a}r, ``Focal loss for
  dense object detection,'' in {\em ICCV}, pp.~2980--2988, 2017.

\bibitem{YOLOF}
Q.~Chen, Y.~Wang, T.~Yang, X.~Zhang, J.~Cheng, and J.~Sun, ``You only look
  one-level feature,'' in {\em CVPR}, pp.~13039--13048, 2021.

\bibitem{YOLOX}
Z.~Ge, S.~Liu, F.~Wang, Z.~Li, and J.~Sun, ``Yolox: Exceeding yolo series in
  2021,'' {\em arXiv preprint arXiv:2107.08430}, 2021.

\bibitem{wang2023yolov7}
C.-Y. Wang, A.~Bochkovskiy, and H.-Y.~M. Liao, ``Yolov7: Trainable
  bag-of-freebies sets new state-of-the-art for real-time object detectors,''
  {\em CVPR}, pp.~7464--7475, 2022.

\bibitem{yolov8}
G.~Jocher, A.~Chaurasia, and J.~Qiu, ``Ultralytics yolov8,'' 2023.

\bibitem{Faster_R-CNN}
S.~Ren, K.~He, R.~Girshick, and J.~Sun, ``Faster r-cnn: Towards real-time
  object detection with region proposal networks,'' {\em TPAMI}, vol.~39,
  no.~06, pp.~1137--1149, 2017.

\bibitem{Libra_R-cnn}
J.~Pang, K.~Chen, J.~Shi, H.~Feng, W.~Ouyang, and D.~Lin, ``Libra {R-CNN:}
  towards balanced learning for object detection,'' in {\em CVPR},
  pp.~821--830, 2019.

\bibitem{Sparse_R-CNN}
P.~Sun, R.~Zhang, Y.~Jiang, T.~Kong, C.~Xu, W.~Zhan, M.~Tomizuka, L.~Li,
  Z.~Yuan, C.~Wang, and P.~Luo, ``{Sparse R-CNN: End-to-End Object Detection
  with Learnable Proposals},'' in {\em CVPR}, pp.~14449--14458, 2021.

\bibitem{yang2023skill}
S.~Yang, L.~Xu, M.~Zhou, X.~Yang, J.~Yang, and Z.~Huang, ``Skill-transferring
  knowledge distillation method,'' {\em IEEE Trans. Circuits Syst.}, vol.~33,
  pp.~6487--6502, 2023.

\bibitem{huang2022feature}
Z.~Huang, S.~Yang, M.~Zhou, Z.~Li, Z.~Gong, and Y.~Chen, ``Feature map
  distillation of thin nets for low-resolution object recognition,'' {\em IEEE
  Transactions on Image Process.}, vol.~31, pp.~1364--1379, 2022.

\bibitem{Deformable_DETR}
X.~Zhu, W.~Su, L.~Lu, B.~Li, X.~Wang, and J.~Dai, ``Deformable detr: Deformable
  transformers for end-to-end object detection,'' in {\em ICLR}, 2021.

\bibitem{Auto-FPN}
A.~Xu, A.~Yao, A.~Li, A.~Liang, and A.~Zhang, ``Auto-fpn: Automatic network
  architecture adaptation for object detection beyond classification,'' in {\em
  ICCV}, pp.~6649--6658, 2019.

\bibitem{FAD}
Y.~Zhong, Z.~Deng, S.~Guo, M.~R. Scott, and W.~Huang, ``Representation sharing
  for fast object detector search and beyond,'' in {\em ECCV}, pp.~471--487,
  2020.

\bibitem{2014Very}
K.~Simonyan and A.~Zisserman, ``Very deep convolutional networks for
  large-scale image recognition,'' {\em arXiv preprint arXiv:1409.1556}, 2014.

\bibitem{2018DARTS}
H.~Liu, K.~Simonyan, and Y.~Yang, ``Darts: Differentiable architecture
  search,'' {\em arXiv preprint arXiv:1806.09055}, 2018.

\bibitem{2014Microsoft}
T.-Y. Lin, M.~Maire, S.~Belongie, J.~Hays, P.~Perona, D.~Ramanan,
  P.~Doll{\'a}r, and C.~L. Zitnick, ``Microsoft coco: Common objects in
  context,'' in {\em ECCV}, pp.~740--755, 2014.

\bibitem{2018MobileNetV2}
M.~Sandler, A.~Howard, M.~Zhu, A.~Zhmoginov, and L.-C. Chen, ``Mobilenetv2:
  Inverted residuals and linear bottlenecks,'' in {\em CVPR}, pp.~4510--4520,
  2018.

\bibitem{RFDNet}
Y.~Zhang, M.~Liu, Y.~Yang, Y.~Guo, and H.~Zhang, ``A unified light framework
  for real-time fault detection of freight train images,'' {\em IEEE Trans.
  Ind. Informat.}, vol.~17, no.~11, pp.~7423--7432, 2021.

\bibitem{resnet-50}
K.~He, X.~Zhang, S.~Ren, and J.~Sun, ``Deep residual learning for image
  recognition,'' in {\em CVPR}, pp.~770--778, 2016.

\bibitem{resnet-101}
K.~He, X.~Zhang, S.~Ren, and J.~Sun, ``Identity mappings in deep residual
  networks,'' in {\em ECCV}, pp.~630--645, 2016.

\bibitem{swin}
Z.~Liu, Y.~Lin, Y.~Cao, H.~Hu, Y.~Wei, Z.~Zhang, S.~Lin, and B.~Guo, ``Swin
  transformer: Hierarchical vision transformer using shifted windows,'' in {\em
  ICCV}, pp.~10012--10022, 2021.

\bibitem{Cspnet}
C.-Y. Wang, H.-Y.~M. Liao, Y.-H. Wu, P.-Y. Chen, J.-W. Hsieh, and I.-H. Yeh,
  ``Cspnet: A new backbone that can enhance learning capability of cnn,'' in
  {\em CVPR}, pp.~390--391, 2020.

\bibitem{CornerNet}
H.~Law, Y.~Teng, O.~Russakovsky, and J.~Deng, ``Cornernet: Detecting objects as
  paired keypoints,'' {\em arXiv preprint arXiv:1808.01244}, 2018.

\end{thebibliography}

\end{document}